\documentclass[runningheads]{llncs}

\usepackage[T1]{fontenc}
\usepackage{fancyvrb}
\usepackage{xcolor}
\usepackage{graphicx}
\usepackage{booktabs}
\usepackage{wrapfig}
\usepackage{hyperref}
\usepackage{color}

\begin{document}
\title{``I understand why I got this grade'': Automatic Short Answer Grading (ASAG) with Feedback}
\titlerunning{``I understand why I got this grade'': ASAG with Feedback}
%
\author{Dishank Aggarwal\thanks{Work done while at IITB}\inst{1}\orcidID{0009-0006-8961-2768} \and
Pritam Sil\inst{2} \orcidID{0000-0002-0992-6892}
\and
Bhaskaran Raman\inst{2} \orcidID{0000-0002-2234-7123}
\and
Pushpak Bhattacharyya \inst{2} \orcidID{0000-0001-5319-5508}
}
\authorrunning{Aggarwal et al.}
\institute{Fujitsu Research of India Private Limited, Bangalore, India \and
Dept. of Computer Science and Engineering, IIT Bombay, Mumbai, India \email{dishank.aggarwal@fujitsu.com \{pritamsil,pb,br\}@cse.iitb.ac.in}}
%

%

\maketitle              
\begin{abstract}
In recent years, there has been a growing interest in using Artificial Intelligence (AI) to automate student assessment in education. Among different types of assessments, summative assessments play a crucial role in evaluating a student's understanding level of a course. Such examinations often involve short-answer questions. However, grading these responses and providing meaningful feedback manually at scale is both time-consuming and labor-intensive. Feedback is particularly important, as it helps students recognize their strengths and areas for improvement. Despite the importance of this task, there is a significant lack of publicly available datasets that support automatic short-answer grading with feedback generation. To address this gap, we introduce Engineering Short Answer Feedback (EngSAF), a dataset designed for automatic short-answer grading with feedback. The dataset covers a diverse range of subjects, questions, and answer patterns from multiple engineering domains and contains $\sim5.8k$ data points. We incorporate feedback into our dataset by leveraging the generative capabilities of state-of-the-art large language models (LLMs) using our Label-Aware Synthetic Feedback Generation (LASFG) strategy.
This paper underscores the importance of enhanced feedback in practical educational settings, outlines dataset annotation and feedback generation processes, conducts a thorough EngSAF analysis, and provides different LLMs-based zero-shot and finetuned baselines for future comparison\footnote{Dataset and code are open-sourced at: \url{https://github.com/dishankaggarwal/EngSAF}}. The best-performing model (Mistral-7B) achieves an overall accuracy of 75.4\% and 58.7\% on unseen answers and unseen question test sets, respectively. Additionally, we demonstrate the efficiency and effectiveness of our ASAG system through its deployment in a real-world end-semester exam at a reputed institute. There, we achieve an output label accuracy of 92.5\% along with feedback correctness and emotional impact scores above 4.5 (out of 5) on human evaluation, thus showcasing its practical viability and potential for broader implementation in educational institutions. 

\end{abstract}

\section{Introduction}
Technology integration in education, particularly artificial intelligence (AI), has resulted in transformative changes that have redefined traditional pedagogical approaches and assessment methodologies. Effective education relies on feedback and explanations provided during assessments to ensure quality learning outcomes \cite{shute2008focus}. In the context of education, educators use different kinds of assessments, including formative, summative, and diagnostic\footnote{https://www.niu.edu/citl/resources/guides/instructional-guide/formative-and-summative-assessment.shtml}, each of which serves a vital role in a student's learning journey. This work deals with summative assessments, which are essential in determining whether the student has achieved the learning goals of a course once it is over. Grading summative assessments often involve grading short answers and essays, which are more complicated in nature due to the flexibility and natural language in the response. Hence, automating the grading process for such questions becomes crucial, especially in countries with extremely high student-to-teacher ratios, as it can significantly reduce instructor's workloads and improve the assessment process. For instance, during the COVID-19 pandemic, educators faced increased stress and exhaustion due to overloaded grading burden \cite{klapproth2020teachers,oliveira2021exploratory}. 

This challenge can be approached as a machine learning problem, where the objective is to grade a student's response based on how similar it is to the reference answers. However, simply assigning a score or label to a learner's response is often inadequate in practical educational contexts \cite{franz2023influence,sree2024empowering}. Ahea et al.~\cite{ahea2016value} highlights the value and effectiveness of feedback in improving student's learning and standardizing teaching in higher education. Although earlier work formulated this problem as the Semantic Textual Similarity (STS) task \cite{agirre2012semeval,agirre2013sem,agirre2014semeval,agirre2015semeval}, recent works acknowledge the importance of feedback generation along with the grading task \cite{filighera2022your}. Hence, together with the grading task and the feedback generation task, researchers have formulated the problem of automatic short answer grading (ASAG) with feedback.

\subsection{Automatic Short Answer Grading (ASAG) with Feedback}
\label{ASAGwithfeedback}
The ASAG with feedback problem is as follows: given a question, a reference answer, and a student's answer, the aim is to assign a label indicating the degree of correctness in the student's answer compared to the reference answer and provide content-focused elaborated feedback/explanation for the same. Note that the degree of correctness is limited to only three labels here, namely correct, partially correct, and incorrect. In this problem, we only focus on questions where the answers are short, varying between a sentence and a short paragraph. This task involves evaluating the alignment between the student's answer and the reference answers which is expressed via the degree of correctness along with proper reasoning as to why that label is assigned.

\begin{figure}[h!tbp] 
  \centering 
  \includegraphics[width=1\textwidth]{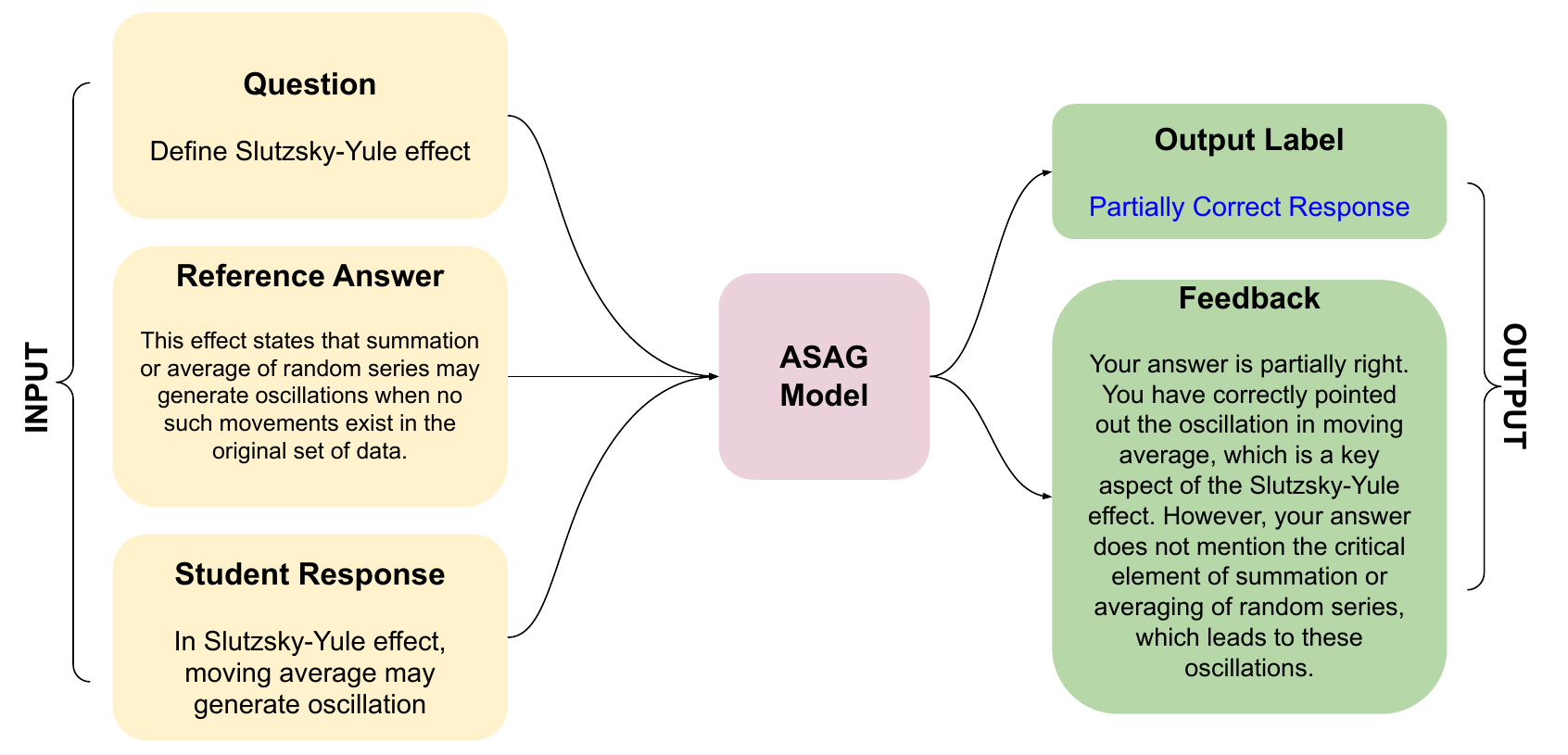} 
  \caption{Illustrative example of the ASAG with Feedback pipeline that classifies a student response as `correct', `partially correct', or `incorrect' and provides feedback.} 
  \label{fig:problem-statement} 
\end{figure}

Figure \ref{fig:problem-statement} illustrates the problem statement, showing the input and output using an example from the EngSAF dataset. %


\subsection{Motivation} 
With recent advancements in state-of-the-art (SOTA) Natural Language Processing  (NLP) techniques, there is an increasing interest in applying such methodologies in the field of education, particularly in short-answer evaluations. The benefits of designing such AI-assisted systems that solve the ASAG with feedback problem are two-fold:
\begin{enumerate}
    \item The learners\footnote{We often use the terms student and learner synonymously. Also, we do the same for the terms graders, teachers, and educators.} benefit by understanding their strengths and shortcomings from the content-elaborated feedback provided by the system.
    \item The educators benefit by having prior feedback on the student's degree of correctness, along with a proper explanation for the same. This can reduce an educator's cognitive load while grading such short answer questions.
\end{enumerate}

However, researchers face challenges when designing such systems due to the lack of public, content-centered, elaborated feedback datasets in different domains. These datasets are crucial for training and developing automated grading systems that can provide feedback. Thus, this calls for a more efficient and effective dataset spanning multiple engineering domains along with proper grades and content-elaborated feedback. This is where we introduce the  Engineering Short Answer Feedback (EngSAF) dataset. \\
 
    \noindent Our contributions are:
    \begin{enumerate}
        \item A Label Aware Synthetic Feedback Generation (LASFG) strategy to convert conventional automatic short answer grading datasets to the ones containing feedback. (Section \ref{LASFG}) 
    
        \item \textbf{EngSAF} dataset containing around \textbf{5.8K} student responses to 119 questions from multiple engineering domains taken from real-life examinations at a reputed institute along with synthetically generated feedback using LASFG strategy explaining the assigned grade for the task of ASAG. To the best of our knowledge, this is the first dataset containing questions and responses from multiple engineering domains. \textit{The overarching goal of this work is to provide helpful feedback to students automatically.} (Section \ref{dataset-section}) 
         \item Benchmark scores on the EngSAF dataset using different Large Language Models (LLMs) for future comparison and research. (Section \ref{results})
         \item Real-world deployment of the best-performing fine-tuned ASAG model in an end-semester exam at a reputed institute. (Section \ref{deployment})
    \end{enumerate}

\section{Related work}
This section provides a comprehensive literature review, beginning with previous research on the ASAG problem, followed by an in-depth survey of existing short-answer grading datasets.

\subsection{Automatic Short Answer Grading (ASAG)}
ASAG is an essential area of research that has garnered significant attention in recent years. Several approaches have been proposed for traditional ASAG techniques, ranging from rule-based methods to more sophisticated machine-learning techniques. One early approach for solving the ASAG problem was based on keyword or pattern matching, where the presence or absence of certain keywords in the student's answer was used to determine its accuracy \cite{mitchell2002towards,sukkarieh2004auto,nielsen2009recognizing}. To overcome these limitations, researchers have developed more sophisticated methods that use NLP-based techniques. One such method is based on Latent Semantic Analysis (LSA), which represents texts as high-dimensional vectors and compares them to the reference answers using cosine similarity \cite{lavoie2020using}. In a related study, the task of ASAG is addressed by incorporating features such as answer length, grammatical correctness, and semantic similarity in comparison to reference answers \cite{sultan2016fast}.

More recently, deep learning models such as convolutional neural networks (CNNs) and recurrent neural networks (RNNs) have been applied to ASAG's task. These models are trained on large amounts of annotated data and can capture the semantic relationships between words in a student's answer and the reference answers \cite{surya2019deep,zhang2022automatic}. Pre-trained Language Models (PLMs), such as BERT \cite{devlin2018bert}, GPT \cite{radford2019language}, RoBERTa \cite{liu2019roberta}, DistillBERT \cite{sanh2019distilbert}, ALBERT \cite{lan2019albert}, and SBERT\cite{reimers2019sentence} have performed exceptionally well for various NLP tasks, are widely used even for the task of ASAG on traditional ASAG datasets \cite{condor2021automatic}. A more recent work by Fateen et al. ~\cite{fateen2024scoresmodularragbasedautomatic} extends this problem by generating feedback while solving the ASAG problem using a RAG-based approach. 
However, a common limitation of all the prior works is that only grades are assigned, and none of the datasets provide a ground truth for the problem content-elaborated feedback generation. 

\subsection{Short Answer Grading Datasets}
Again, when it comes to short answer grading datasets, researchers have carefully curated several publicly accessible datasets to facilitate research and benchmarking of different solutions on the ASAG task.
In 2009, Mohler et al.~\cite{mohler2009text} published a short answer dataset on a data structures course that contained 630 records. In 2011, the same authors~ \cite{mohler2011learning} published an extended dataset on the same course that contained 2273 records and named it the Texas Extended dataset. Apart from this, Nielsen et al.~\cite{nielsen2008annotating} created the Beetle corpus, which contains fine-grained annotations of entailment relationships in the context of student answers to science assessment questions.  Another dataset that is publicly available on Kaggle and could be used for the task of ASAG is provided by Hewlett Foundation named ASAP-AES\footnote{\url{https://www.kaggle.com/c/asap-aes/data}} (Automated Assessment Prize Competition for Essay Scoring).

However, as mentioned above, all datasets are conventional ASAG datasets that solely consist of grades or scores. 
Our extensive literature survey includes only one short-answer grading dataset with content-focused elaborated feedback, i.e., \cite{filighera2022your}, which introduces an inaugural dataset for this problem and comprises bilingual responses in English and German. Human annotations were meticulously collected and refined to uphold the quality of the feedback. However, this dataset is constrained by its limited number of student responses ($\sim2k$) in English and its exclusive focus on questions from a single domain, specifically computer network queries, thereby lacking diversity across different engineering domains. To address these limitations, we curate the EngSAF dataset, which contains $\sim5.8k$ data points and questions from multiple engineering domains.

\section{Engineering Short Answer Feedback (EngSAF) Dataset}
\label{dataset-section}

This dataset contains 119 questions drawn from different undergraduate and graduate engineering courses, accompanied by approximately 5.8k student responses. An instructor provided correct/reference answer also accompanies each question. These questions and answers have been taken from real-life examinations at a reputed institute, thus covering a diverse range of 25 courses. The questions and responses span across multiple domains, including image processing, water quality management, and operating systems, to name a few. The dataset has been curated using a well-known e-learning platform that has been extensively used and developed at a reputed institute. 
Based on the instructor's assigned marks, each response has been categorized as ``\textit{correct},'' ``\textit{partially correct},'' or ``\textit{incorrect}.'' 

\begin{table}[h]
\centering
\caption{An example showing a question, reference answer, and three student answers (Student\#1, Student\#2, and Student\#3) alongside their corresponding labels and synthetically generated Feedback/Explanation for the assigned label from the EngSAF Dataset }
\begin{tabular}{p{0.25\linewidth}  p{0.65\linewidth} }
\toprule
\textbf{Question} & What is the difference between basin order and channel order?\\
\midrule
\textbf{Reference       Answer} &Basin order is highest order of any stream in that basin whereas channel order is order of stream
which denotes that in what order of streams has joined the channel."\\
\midrule
\textbf{Student Answer 1 }& Highest order channel is the basin order whereas channel order is the order of channel from tributaries to reaches to main river stream.\\
\textbf{Label}&\textcolor{green}{\textbf{2 (Correct response)}}\\
\textbf{Feedback}& Excellent! You have a clear understanding of the distinction between basin order and channel order.\\
\midrule
\textbf{Student Answer 2} &Channel order reflects to the number of streams coming together to form a channel.\\
\textbf{Label}&\textcolor{blue}{\textbf{1 (Partially Correct response)}}\\
\textbf{Feedback}&Your answer includes a part of the distinction. Channel order indeed indicates the number of streams joining together to make a channel, but the difference between basin order and channel order is not mentioned.\\
\midrule
\textbf{Student Answer 3} &Channel order is the order of the highest order streams. For example, two first order streams (or more) will make a second order stream and similarly for highest orders.\\
\textbf{Label}&\textcolor{red}{\textbf{0 (Incorrect response)}}\\
\textbf{Feedback}& The student answer confuses basin order with channel order. Basin order refers to the highest order of streams within a basin, while channel order refers to the order of streams based on the sequence of junctions.\\
\bottomrule 
\end{tabular}

\label{tab:Eng-SAF_example}
\end{table}

Table \ref{tab:Eng-SAF_example} provides examples from the EngSAF dataset showcasing a question, its corresponding reference answer, and three different student responses alongside their associated output labels and synthetically generated feedback. 
The dataset is applicable for both traditional automatic short answer grading and the generation of elaborated feedback.

\subsection{Dataset Construction}

The EngSAF dataset contains questions and student responses from different undergraduate and graduate engineering courses from a reputed institute. The data was collected using a well-known e-learning platform that has been developed and is extensively used at this institute. Using the e-learning platform, the instructor provided a correct answer/ reference answer to each question.  Additionally, the instructors and teaching assistants (TAs) have assigned marks to each student's answer from their respective courses. The student's responses and grades were then obtained from the e-learning platform. The quality assessment of the proposed dataset is covered in subsection \ref{QA_Estimation}. 

\begin{table}[h]
    \centering
            \caption{Labelling scheme used for the annotation. Here, $x$ is the marks given by the instructor/TA, and $k$ is the maximum marks for that question.}
    \begin{tabular}{ c c } 
    \toprule
    \textbf{Condition}& \textbf{Output Label}\\
    \midrule
    $x=k$& 2 (Correct Response)\\
    $0<x<k$& 1 (Partially Correct Response)\\
    $x=0$& 0 (Incorrect Response)\\
    \bottomrule
    \end{tabular}

    \label{tab:Annotation details}
\end{table}
Each data point includes a question, student answer, and reference answer, along with a label indicating the level of correctness of the student's response and detailed, content-specific feedback. If the maximum mark for a question is $k$ and a student's answer is graded $x$ marks by the instructor, then the output label is annotated according to the scheme in Table \ref{tab:Annotation details}.

\subsection{Label Aware Synthetic Feedback Generation (LASFG)}
\label{LASFG}
Apart from the question, the student's answer, and the reference answer, each data point contains a content-elaborated feedback part. To obtain this, we leverage the advanced language generative and reasoning capabilities of  SOTA LLMs like Gemini\footnote{\url{https://gemini.google.com/}} and ChatGPT.
The approach involves utilizing Gemini's ability to comprehend input prompts consisting of a question, a student's answer, a reference answer, and the corresponding grading label provided by an instructor or teaching assistant to generate content-focused elaborated feedback.  The generated feedback covers the reasoning or explanation of the Gold output label. Based on initial observations of the generated feedback quality, we use \Verb#gemini-1.5-flash# model for feedback generation. To ensure the quality of the generated feedback, we performed quality estimation as described in Section \ref{QA_Estimation}.
\subsection{Corpus Statistics}

Following Dzikovska et al.'s approach~\cite{dzikovska2013semeval}, we partitioned the data into training sets, comprising 70\% of the dataset, as well as unseen answers (16\%) and unseen questions (14\%) for test sets as shown in Table \ref{tab:split}. The test split with unseen answers (\textbf{\textit{UA}}) includes fresh responses to the questions used in training. In contrast, the test split with unseen questions (\textbf{\textit{UQ}}) comprises entirely new questions for testing the model's ability to generalize to new questions without prior exposure. The average text length (in tokens) for different fields in our dataset is as follows: Questions have an average length of \textbf{17.50} tokens, student answers average \textbf{25.85} tokens, reference answers are \textbf{26.47} tokens long on average, and feedback is the longest, with an average of \textbf{41.95} tokens.

\begin{table}[h]

    \centering
    \caption{Distribution of gold label outputs across train and test split on EngSAF dataset. The test set is further subdivided into Unseen Answers (UA) and Unseen Questions (UQ). }
    \footnotesize
    \begin{tabular}{ c c c c | c } 
    \toprule
    \textbf{Label}& \textbf{Train}& \textbf{UA}& \textbf{UQ} & \textbf{Total}\\
    \midrule
    Correct  &1716&403&321&2440\\
    Partially Correct  &1412&344&278&2034\\
    Incorrect  &941&233&166&1340\\
    
    \midrule
    Total& 4069 &980&765&5814\\
    \bottomrule
    \end{tabular}
        
    \label{tab:split}
\end{table}
\subsection{Quality Estimation}
\label{QA_Estimation}

To show the reliability and credibility of our dataset and synthetically generated feedback, we randomly sampled \textbf{300 data points} and equally distributed them across the output label. This sampled data is a good representation of the EngSAF dataset. Each synthetically generated feedback was evaluated on three aspects by three proficient English-speaking annotators, all Master's students. They were provided with detailed annotation guidelines and examples to ensure consistency.

Each aspect is scored on a scale (1-5), with a high score indicating a better response. Each feedback is analyzed on the following aspects.

\begin{enumerate}
    \item \textbf{Fluency and Grammatical Correctness (FGC)}: This aspect checks if the feedback is fluent and grammatically correct in English.
    \item \textbf{Feedback Correctness/Accuracy (FC)}: This aspect assesses the overall quality of the generated feedback regarding content, relevance, quality, and explanation for the assigned grade.
    \item \textbf{Emotional Impact (EI)}: This aspect assesses how feedback affects the learner's emotional state. Annotators are tasked with rating whether the feedback avoids triggering negative emotions or impacts by refraining from using words such as ``fail'' that may evoke feelings of discouragement or distress in the learner.
\end{enumerate}

\begin{table}[h!tbp]
    \centering
    
    \caption{Average human annotation scores across Fluency \& Grammatical Correctness (FGC), Feedback Correctness/Accuracy (FC), and Emotional Impact (EI)}
    \footnotesize
    \begin{tabular}{ c |c } 
    \toprule
    \textbf{Aspect}& \textbf{ Avg. Score}\\
    \midrule
    Fluency \& Grammatical Correctness (FGC) & 4.73 \\
    Feedback Correctness/Accuracy (FC) & 4.55\\
    Emotional Impact (EI) & 4.61\\
    \bottomrule
    \end{tabular}
    \label{tab:human_eval}
\end{table}

Three human annotators evaluate the correctness of each output label for sampled data points, achieving an accuracy of \textbf{98\%} and pair-wise average Cohen's Kappa ($\kappa$) score of \textbf{0.65} (substantial agreement), showcasing the high reliability of the assigned output label.
Table \ref{tab:human_eval} shows the average score over all the designed aspects to measure the reliability of the proposed dataset. Human evaluation across various designed aspects consistently yields an average score greater than \textbf{4.5} out of 5, which underscores the reliability and quality of the EngSAF dataset. 

As part of our ablation studies, we have also performed an LLM-based evaluation by following the LLM-as-a-judge paradigm~\cite{ZHengLLMasaJudge}. We conducted such an evaluation using the Gemini model to assess the quality of the generated feedback. Specifically, we instructed Gemini to evaluate feedback quality on the EngSAF test set, and we obtained average scores of \textbf{4.23} for feedback quality on unseen answers (UA) and \textbf{3.91} on unseen questions (UQ) out of 5. These scores reflect the model's assessment of the feedback quality and provide an additional layer of validation to the reliability and quality of the dataset.
    \section{Experiments}
    \label{Experiments}

Following Filighera et at.'s \cite{filighera2022your} methodology, our primary objective is to establish a baseline for our EngSAF dataset. We also assess the impact of incorporating questions on the generated feedback and the assigned labels, by performing experiments in two different settings, namely \textbf{with\_question} and \textbf{without\_question}. Traditionally, in Automatic Short Answer Grading (ASAG), assessments have focused solely on evaluating reference answers and student responses. However, Lv et al.~\cite{lv2021exploring} challenge this convention by demonstrating that integrating questions into the evaluation process improves the performance of traditional ASAG tasks.

\subsection{Experimental Setting}

To establish baselines for EngSAF, we have conducted experiments in two settings.
\begin{enumerate}
    \item \textbf{Fine-tuning Large Language Models (LLMs)}: We use the Llama-2 model \cite{touvron2023llama}, Llama-3.1 model \cite{dubey2024llama}, Mistral 7b model\footnote{\url{https://mistral.ai/}}, which are fine-tuned to predict the output label for student responses, categorizing them as correct, partially correct, or incorrect and jointly providing feedback explaining the assigned output label. Furthermore, we have conducted this experiment using two distinct methodologies.

\begin{enumerate}
    \item \textbf{Without Question}: Student answer and Reference answer are passed as input.
    \item \textbf{With Question}: Question, Student answer, and Reference answer are passed as input.
\end{enumerate}

 The models used in this experiment are \Verb#llama-3.1-8B-Instruct-chat#, \Verb#llama-2-13B# and \Verb#Mistral-7B-Instruct-v0.1#.

\item \textbf{Zero-shot experimentation}: For this experiment, we prompt ChatGPT\footnote{\url{https://openai.com/}} and DeepSeek \cite{guo2025deepseek} to assign an output label along with feedback to each student's response by evaluating its correctness compared to the reference answer for the ASAG task. The following models are used: \Verb#gpt-4o# and \Verb#DeepSeek-R1#
\end{enumerate}
 
All experiments are performed on 2 \textbf{NVIDIA A100-SXM GPU} with \textbf{80 GB} of memory. Fine-tuning takes around 2 hours per epoch for training.

\section{Results}
\label{results}
\begin{table}[h!tbp]

\centering
\caption{Fine-tuning and zero-shot experiment results on the EngSAG unseen answers and unseen
questions test splits. w\_quest models additionally received the questions as
input, while wo\_quest did not. Please note that the text similarity measures, accuracy, and F1 scores are in percent. Zero-shot experiments were conducted solely on the unseen answers test set, as both unseen questions and answers are identical in
zero-shot setting.}
\resizebox{\columnwidth}{!}{\begin{tabular}{c|cccccc|cccccc}
\toprule
& \multicolumn{6}{c}{Unseen Answers} & \multicolumn{6}{c}{Unseen Questions} \\
Model& Acc. &F1 & BLEU & MET. & ROU. & BERT & Acc. & F1 & BLEU & MET. & ROU.& BERT\\
\midrule
Majority& 43.3 &26.1& 1.2 & 8.6 & 12.7 & 16.2 & 40.5 & 23.4 & 0.1 & 8.64 & 2.78 & 12.32\\
\midrule
\multicolumn{13}{c}{Fine-tuning experiments}\\
$Llama-2_{wo\_quest}$& 74.4&73.0& 13.3 &34.1 & 16.8& 31.9 & 55.6 & 53.6 & 12.5 & 31.4 & 16.2 & 28.9\\
$Llama-2_{w\_quest}$& 73.9&73.7& 11.7 &35.9 & 16.9& 35.1& 56.3 & 54.9 & 9.2 & 31.4 & 13.9 & 32.6\\
$Llama-3.1_{wo\_quest}$& 70.1 & 69.4 & 13.2 & 34.7 & 17.7 & \textbf{42.4} & 53.5 & 52.0 & 10.8 & 30.6 & 15.1 & 40.1\\
$Llama-3.1_{w\_quest}$& 72.4 & 72.9 & 13.6 & 35.7 & 17.6 & 40.4 & 54.9 & 56.1 & 13.4 & 34.4 & 17.5 & 39.1\\
$Mistral_{wo\_quest}$&72.8 & 73.1& 11.3 &33.3 & 14.82& 37.4& 54.7 & 55.4 & 11.3 & 32.0 & 15.2 & 36.2\\
$Mistral_{w\_quest}$& \textbf{75.4}&\textbf{75.7}& \textbf{13.9} &\textbf{38.3} & \textbf{19.5}& 41.6& \textbf{58.7} & \textbf{57.9} & \textbf{14.9} & \textbf{36.7} & \textbf{19.7} & \textbf{40.9}\\

\midrule
\multicolumn{13}{c}{Zero-shot experiments}\\

ChatGPT-4o& 48.9   &46.9 & 3.64 & 22.3 & 5.2 & 20.8 & - & - & - & - & - & -\\
DeepSeek-R1 & 47.4 & 46.7 & 4.1 & 23.6 & 6.5 & 18.0 & - & - & - & - & - & -\\
\bottomrule
\end{tabular}}

\label{tab:baseline_results}
\end{table}

Table \ref{tab:baseline_results} shows a majority baseline, fine-tuning experiments, and zero-shot experiments results. The majority baseline contains the most occurring label and feedback from the EngSAF train set. The most common label is ``correct response,'' and the most common feedback is ``Well done! You have answered the question correctly, covering all the required aspects.''.

\subsection{Analysis}
\label{sec:Analysis}
In this section, we delve into the insights drawn from the performance results of experiments on the EngSAF dataset for the ASAG task. Following Filighera et at.’s \cite{filighera2022your} evaluation methodology, we measure the accuracy and macro-averaged F1 for output labels and the ROUGE-2 \cite{post2018call}, SACREBLEU\footnote{\url{https://pypi.org/project/sacrebleu/}}, METEOR \cite{banerjee2005meteor} and BERTScore \cite{zhang2019bertscore} for the feedback part. For all metrics, a higher score indicates better results.

As per Table \ref{tab:baseline_results}, Mistral-7b significantly outperforms the majority baseline in both output label and feedback metrics.
However, we observe a significant performance gap between unseen answers and unseen questions, with accuracy dropping by approximately\textbf{ 23\% }for unseen questions and by\textbf{ 4\%} in BERTScore, suggesting the necessity for new evaluation metrics that comprehensively assess text on the context level instead of the lexical level. The above two observations highlight the challenge that even fine-tuned models face when generalizing to new questions or domains (UQ). In the case of unseen answers (UA), other responses to the same question are present in the training data, making generalization easier compared to entirely new questions. A similar pattern can be observed for other fine-tuned models. Notably, including questions in the input significantly improves accuracy, F1 score, and BERTScore. For example, in our experiments with Llama-2 and Mistral-7B, incorporating the question led to a \textbf{10\%} improvement in BERTScore for unseen answers and a \textbf{12\%} improvement for unseen questions.

Our zero-shot experiments with ChatGPT and DeepSeek-R1 yield an accuracy of \textbf{48.9\%} and \textbf{47.4\%}, respectively, falling below the fine-tuning baselines, highlighting the complexity of the ASAG task. We observe a similar performance gap in the feedback evaluation part, suggesting that fine-tuning plays a crucial role in capturing complex grading patterns. Another contributing factor to this performance disparity of zero-shot experiments in comparison to fine-tuned experiments could be the inherent bias present in teacher grading, which models can only learn when trained on such data. 
\section{Real-World Deployment}
\label{deployment}

For real-world deployment, we integrated our best-performing fine-tuned ASAG model for automatic evaluation of an end-semester exam of a course at a reputed institute. The end-semester examination included two short-answer questions, each accompanied by the instructor's correct/reference answer. We randomly sampled 25 student answers for each question and used our fine-tuned $Mistral_{w\_quest}$ model to predict the output label and feedback/ explanation for the predicted output. To ascertain the quality of the generated output, we employed \textbf{three} Subject Matter Experts (SMEs) to evaluate the predicted output and feedback. The SMEs were PhD students who were teaching assistants (TAs) of that course. Each predicted output label was evaluated based on its correctness, where the subject matter expert checks whether the predicted output label is correct or not. Each predicted feedback is analyzed in terms of \textbf{Feedback Quality/Correctness (FC)} and \textbf{Emotional Impact (EI)} as discussed in the Quality Estimate section \ref{QA_Estimation}. Each aspect is scored on a \textbf{scale (1-5)}, with a high score indicating a better response.

\begin{table}
    \centering
\caption{Subject Matter Experts (SME) scores evaluating the real-world deployment across label accuracy, Feedback Quality/ Correctness (FC), and Emotional Impact (EI)}
    
    \footnotesize
    
    \begin{tabular}{ c |c } 
    \toprule
    \textbf{Aspect}& \textbf{ Avg. Score}\\
    \midrule

    Output Label Accuracy & 92.5\%\\
    Feedback Quality/Correctness (FC) & 4.5\\
    Emotional Impact (EI)& 4.9\\
    \bottomrule
    \end{tabular}
    
    \label{tab:ET623_eval}
    
\end{table}

Table \ref{tab:ET623_eval}  presents the average scores across all the aspects used to measure the reliability of our ASAG model. Upon examining the table, it is evident that the subject matter evaluation scores are greater than \textbf{4.5} for both FC and the EI aspect. This demonstrates the reliability and effectiveness of the ASAG model in real-world scenarios. Additionally, the model's predicted output label achieved an accuracy of \textbf{92.5\%}, further showcasing its performance and reliability. By providing accurate labels and feedback at scale, our approach helps reduce the manual workload for teachers and instructors, making the grading process more efficient and scalable.
\section{Conclusion}
\label{summary}

This work introduces a novel and extensive dataset for automatic short answer grading (ASAG) across multiple engineering domains. Addressing the need for a standardized evaluation platform, our dataset encompasses diverse subjects and a wide range of short-answer responses. We meticulously curated and annotated the dataset to ensure its quality, consistency, and applicability to real-world grading scenarios. We propose Label-Aware Synthetic Feedback Generation (LASFG)—a method that enables the transformation of any traditional ASAG dataset into one that incorporates meaningful synthetic feedback. Additionally, we benchmark our dataset using fine-tuned large language models (LLMs) and conduct zero-shot evaluations, providing a comprehensive assessment of its effectiveness.

Furthermore, we deployed our ASAG system in a real-world course exam at a prestigious institution, showcasing its practical utility in educational settings. Human evaluation showcases the effectiveness of the ASAG system by achieving scores more than \textbf{4.5 out of 5} on Feedback Quality/Correctness (FC) and Emotional Impact (EI). Additionally, the model demonstrates its ability to correctly detect an answer with an accuracy of \textbf{92.5\%}. This work represents a significant milestone in educational technology, offering educators and researchers a valuable tool to drive innovation in automated assessment methods.

Future extensions of this work could focus on expanding the dataset to include more complex short-answer responses, enabling broader grading scenarios. Additionally, incorporating the evaluation of diagrams and images within student responses could further enhance the system’s capabilities, paving the way for more comprehensive automated assessment solutions.

\bibliographystyle{splncs04}
\bibliography{AIED_EngSAF_Dishank}

\end{document}